\documentclass{bmvc2k}

\title{Visually Aligned Word Embeddings for Improving Zero-shot Learning}

\addauthor{Ruizhi Qiao}{ruizhi.qiao@adelaide.edu.au}{1}
\addauthor{Lingqiao Liu}{lingqiao.liu@adelaide.edu.au}{1}
\addauthor{Chunhua Shen}{chunhua.shen@adelaide.edu.au}{1}
\addauthor{Anton van den Hengel}{anton.vandenhengel@adelaide.edu.au}{1}

\addinstitution{
 School of Computer Science, University of Adelaide,
 Australia
}

\runninghead{Qiao {\it et al.}}{Visually Aligned Word Embeddings. Appearing in Proc. British Mach. Vis. Conf. 2017}

\def\argmax{\operatorname*{argmax\,}}

\def\Real{\mathbb{R}}
\def\Natural{\mathbb{N}}

\def\bv{ {\bf v } }

\def\bs{ {\bf s } }

\def\calX{ {\cal X } }

\def\calW{ {\cal W } }

\def\calT { {\cal T} }

\def\calY { {\cal Y} }

\usepackage{amsmath}
\usepackage{amssymb}
\usepackage{epsfig}
\usepackage{graphicx}
\usepackage{algorithm2e}
\let\savedalgorithm\algorithm
\let\savedendalgorithm\endalgorithm

\usepackage{algorithm}
\usepackage{footnote}

\begin{document}

\maketitle

\begin{abstract}
Zero-shot learning (ZSL) highly depends on a good semantic embedding to connect the seen and unseen classes. Recently, distributed word embeddings (DWE) pre-trained from large text corpus have become a popular choice to draw such a connection. Compared with human defined attributes, DWEs are more scalable and easier to obtain. However, they are designed to reflect semantic similarity rather than visual similarity and thus using them in ZSL often leads to inferior performance. To overcome this visual-semantic discrepancy, this work proposes an objective function to re-align the distributed word embeddings with visual information by learning a neural network to map it into a new representation called visually aligned word embedding (VAWE). Thus the neighbourhood structure of VAWEs becomes similar to that in the visual domain.  Note that in this work we do not design a ZSL method that projects the visual features and semantic embeddings onto a shared space but just impose a requirement on the structure of the mapped word embeddings. This strategy allows the learned VAWE to generalize to various ZSL methods and visual features. As evaluated via four state-of-the-art ZSL methods on four benchmark datasets, the VAWE exhibit consistent performance improvement.
\end{abstract}

\section{Introduction}
\label{sec:intro}
Zero-shot learning (ZSL) aims at recognizing objects of categories that are not available at the training stage. Its basic idea is to transfer visual knowledge learned from seen categories to the unseen categories through the connection made by the semantic embeddings of classes. Attribute \cite{Farhadi09describingobjects} is the first kind of semantic embedding utilized for ZSL and remains the best choice for achieving the state-of-the-art performance of ZSL ~\cite{Akata15output,Zhang2015ICCV}. Its good performance, however, is obtained at the cost of extensive human labour to label these attributes.

\begin{figure}[ht]
\begin{center}\label{fig:overview}
   \includegraphics[width=0.6\linewidth]{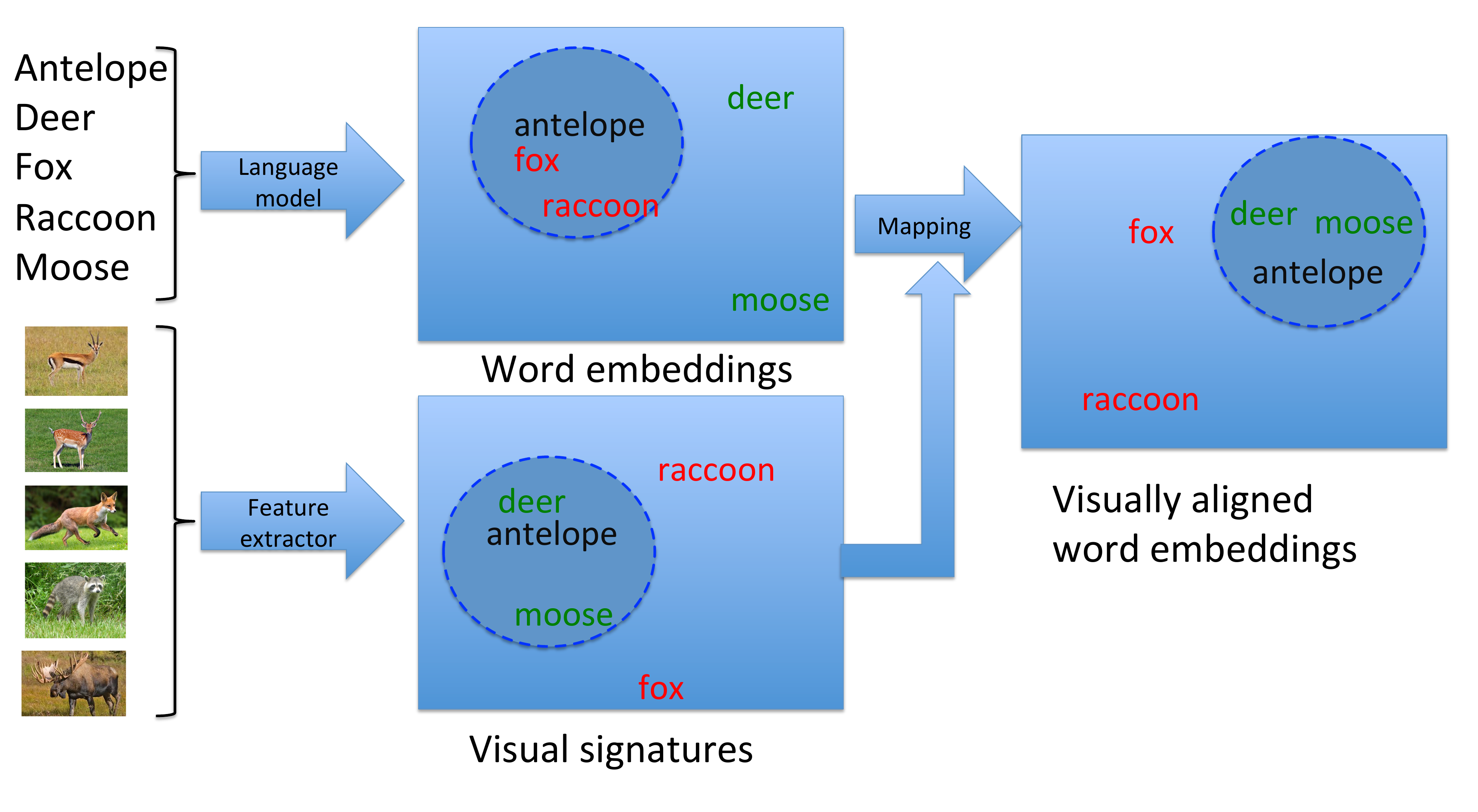}
\end{center}
   \caption{The key idea of our approach. Given class names and visual features of the seen classes, we extract the word embeddings from a pre-trained language model and obtain the visual signatures that summarize the appearances of the seen classes. The word embeddings are mapped to a new space where the neighbourhood structure of the mapped embeddings are enforced to be consistent with their visual domain counterparts. During the inference stage, the VAWEs and visual features of seen classes are used to train the ZSL model. Then VAWEs of unseen classes are fed to the trained ZSL model for zero-shot prediction.}
\label{fig:overview}
\end{figure}

Recently, several works have explored to use distributed word embeddings (DWE)  \cite{word2vec,glove} as the alternative to attributes in zero-shot learning \cite{Frome2013NIPS_devise,NorouziConse13}. In contrast to human annotated attributes, DWEs are learned from a large-scale text corpus in an unsupervised fashion, which requires little or no human labour to collect. However, the training process of DWEs does not involve visual information and thus they only capture the semantic relationship between different classes. In practice, the semantic similarity does not necessarily correspond to the visual similarity and this visual-semantic discrepancy may lead to the inferior performance of ZSL. In fact, it has been shown that when applied to the same ZSL approach, DWE is always outperformed by attribute \cite{Akata15output,SynC2016,LatEm2016}. To reduce the visual-semantic discrepancy, a popular way in ZSL is to map the semantic embeddings and visual features into a shared space \cite{Frome2013NIPS_devise,Romera2015ZSL,LatEm2016,Zhang2016CVPR, LongBMVC16} to make these two domains comparable. However, when a large visual-semantic discrepancy exists, finding such a mapping can be difficult.%

Different to existing work, the method proposed in this paper directly learns a neural network to map the semantic embedding to a space in which the mapped semantic embeddings preserves a similar neighbourhood structure as their visual counterparts. In other words, we do not require the mapped semantic embeddings to be comparable to visual features but only impose constraints on their structure. This gives more freedom in learning the mapping function, and this could potentially enhance its generalizability. Moreover, since our approach is not tied to a particular zero-shot learning method, the learned mapping can be applied to any zero-shot learning algorithm.

Three contributions are made in this work. First, we experimentally demonstrate that the inferior ZSL performance of DWE is caused by the discrepancy of visual features and semantic embeddings. Second, to overcome this issue, we propose the visually aligned word embeddings (VAWE) which preserve similar neighbourhood structure with that in the visual domain. Third, we show that VAWE has improved the word embedding based ZSL methods to state-of-the-art performance and is potentially generalizable to any type of ZSL method.

\section{Related works}
\label{sec: rel_work}
{\bf Zero-shot learning and semantic embedding:} Zero-shot learning was firstly made possible by attributes\cite{Lampert09unseen,Farhadi09describingobjects}, which describe the visual appearance of the concept or instance by assigning labelled visual properties to it, and they are easily transferable from seen to unseen classes. Distributed word embeddings, most notably word2vec \cite{word2vec} and GloVe \cite{glove}, are recently explored \cite{SocherNIPS2013,Frome2013NIPS_devise,NorouziConse13} as a promising alternative semantic embedding towards fully automatic zero-shot learning since their unsupervised training process does not involve any human intervention. ZSL approaches learn a connection between visual and semantic domains either by directly mapping visual features to semantic space \cite{SocherNIPS2013,NorouziConse13,fu2016semi} or projecting both visual and semantic embeddings into a common space \cite{Akata13label,Frome2013NIPS_devise,Romera2015ZSL,Akata15output,LatEm2016,Zhang2016CVPR, LongBMVC16, SaumyaBMVC15, LiICCV15}. It should be noted that specically in \cite{LongBMVC16, LongCVPR17}, similar issues like visual-semantic ambiguity or visual-semantic structure preservation are proposed and attribute based ZSL methods are designed to deal with them. Although our work shares a common goal with \cite{LongBMVC16} and \cite{LongCVPR17}, VAWE is learned in the semantic domain only which serves as a general tool for any word embedding based ZSL methods. In other words, we are {\bf NOT} proposing a particular ZSL method, and VAWE can be regarded as a meta-method for improving existing ZSL methods.

{\bf Word embedding with visual information:} As distributed word embedding is limited to pure textual representation, a few works have proposed to improve it with visual cues. Visual word2vec \cite{Kottur_2016_CVPR} is trained by adding abstract scenes to context.  In \cite{Lazaridou_combine_vs2015}, the language model learns to predict visual representations jointly with the linguistic features. Our work is different from those two works in two aspects: 1) our target is to learn a mapping function which can be generalized to words in unseen classes while the above works try to learn an embedding for the words in the training set. 2) the objective of our method is to encourage a certain neighbourhood structure of the mapped word embedding rather than applying the context prediction objective across visual and semantic domains as in \cite{Kottur_2016_CVPR,Lazaridou_combine_vs2015}.

\section{Background and motivation}
\label{sec: bg}

Assume a set of class labels $\calW_s$ and $\calW_u$ for images from seen and unseen classes, where $\calW_s \cap \calW_u = \varnothing$. Most zero-shot learning approaches can be summarised by the general form

\begin{align}\label{eq:formulation}
 s(x) = \argmax_{y \in \calY} {F(x, \phi(y))},
\end{align}
where $F(x, \phi(y))$ measures compatibility score of the visual feature $x$ and a semantic embedding $\phi(\cdot)$ of class $y$. During the training phase, where $\calY = \calW_s$, $F(\cdot, \cdot)$ is learned to measure the compatibility between $x$ and $\phi(y)$. During the testing phase, the learned $F(x, \phi(y))$ is applied to measure the compatibility between novel classes $y \in \calW_u$ and testing visual samples  $x \in \calX_{unseen}$ for zero-shot prediction.

This formulation indicates that $\phi(y)$ is an important factor of ZSL. It is desirable that the relationship among $\phi(y)$ retains consistency with the relationship among their visual features, that is, $\phi(y)$ of the visually similar classes remains to be similar and vice versa. Human defined attributes are empirically proven to be qualified $\phi(y)$ because the annotators implicitly infuses the attribute vectors with visual information based on their knowledge and experience of the concepts\footnote{In this work, we use ``concept'' and ``word'' interchangeably to denote category names.}. However, this is not always the case for semantic embeddings learned from pure text sources, which are trained to maintain the semantic relation of concepts from large text corpora.  For example, the concepts ``violin'' and ``piano'' are strongly related in the semantic sense even though their appearances are completely different.

To investigate how visual-semantic consistency affects ZSL performance, we conduct a preliminary experiment on AwA dataset\footnote{See Section~\ref{sec: exp} and the supplementary material for the details of this dataset and the visual features}. We use the state-of-the-art ESZSL \cite{Romera2015ZSL} method to measure the ZSL performance and the average neighbourhood overlap to measure visual-semantic consistency. To calculate the latter, we measure the visual distance between two classes as the average distance between all pairs of visual features within those two classes and this is also equivalent to calculating the distance between their mean feature vectors.That is to say,the visual distance between two classes $i$ and $j$ are
\begin{align}\label{eq:vdist}
 D_{i,j} = \|{\bf f}_i - {\bf f}_j \|_2
\end{align}

where ${\bf f}_i$ is the mean feature vectors for each class and $\|\cdot \|_2$ is the L-2 norm. Likewise, the semantic distance between two classes can be calculated in the same manner by replacing ${\bf f}_i$ and ${\bf f}_j$ with the semantic embeddings of classes $i$ and $j$.

We define $N_v(i, K)$ and $N_s(i, K)$ as the sets that includes the $K$ most similar classes to class $i$ in visual and semantic domains respectively.
Then for each class $i$, we calculate its top-$K$ nearest classes in visual domain using \eqref{eq:vdist} and put them in $N_v(i, K)$. Similarly, we calculate the top-$K$ nearest classes of $i$ in the semantic domain and put them in $N_s(i, K)$.

Four types of semantic embeddings: word2vec, GloVe \footnote{See Section~\ref{sec: exp} and the supplementary material for the training details of the two word embeddings}, binary attribute (presence/absence of the attribute for a class) and continuous attribute (strength of association to a class) are tested. The average neighbourhood overlap is defined in \eqref{eq:const} as the average number of shared neighbours (out of $K$=10 nearest neighbours in this case) for all $C$ classes in semantic and visual domains. A value closer to $10$ indicates that the embedding is more consistent with the visual domain.

\begin{align}\label{eq:const}
 Consistancy = \sum_{i=1, \cdots, C} {|N_v(i, K) \cap N_s(i, K) |} / C
\end{align}

\begin{table}[ht]
\begin{center}
\begin{tabular}{|l|c|c|c|}
\hline
Method &Embedding &Consistency  &Accuracy\\
\hline\hline
ESZSL   &word2vec &2.88 &58.12 \\
ESZSL  &GloVe &2.84  &59.72\\
ESZSL  &binary attribute &4.80   &62.85\\
ESZSL  &continuous attribute &5.66   &75.12\\
ESZSL  &visual feature mean &10.00 &86.34 \\

\hline
\end{tabular}
\end{center}
\caption{Preliminary experiment: ZSL accuracies of ESZSL on AwA dataset with different semantic embeddings. The visual feature mean summaries the visual appearance of each seen or unseen class. }
\label{tab: pre_test}
\end{table}

The results in Table~\ref{tab: pre_test} demonstrate that semantic embeddings with more consistent visual-semantic neighbourhood structure clearly produce better ZSL performance. Motivated by that, in this paper we propose to map the semantic word embedding into a new space in which the neighbourhood structure of the mapped embeddings becomes consistent with their visual domain counterparts. Hereafter, we call the mapped word embedding visually aligned word embedding (VAWE) since the mapped word embedding is re-aligned with visual information in comparison with its unmapped counterpart.

\section{Approach}

{\bf Notations:} Before elaborating our approach, we formally define the notations as follows. For each visual category $i$, we denote its semantic word embedding as $\bs_i \in {\Real}^{d^s}$ and its visual signature as $\bv_i \in {\Real}^{d^v}$, where $d^v$ and $d^s$ are the dimensionality of the visual and semantic space, respectively. The visual signature will be used to define the neighbourhood structure in the visual domain. In the main body of this paper, we use the mean vector of the visual features in the $i$th category as its visual signature. Certainly, this is merely one way to define the visual neighbourhood structure, and our method also applies to other alternative definitions.
The to-be-learned mapping function (neural network) is represented by $f_\Theta(\cdot)$ \footnote{For simplicity, we omit the parameter $\Theta$ in later notations.}, where $\Theta$ is the model parameters. This function will be learned on the seen classes and is expected to generalize to unseen classes. In this way, we can apply the VAWE to any zero-shot learning methods. We use the notation $i^*\in \calW_u$ and $\bs_i^*$ to denote an unseen class and its semantic embedding respectively.

\subsection{Visually aligned word embedding}
\label{sec: structure}
To learn $f(\cdot)$, we design an objective function to encourage that $f(\bs_i)$ and $\bv_i$ share similar neighbours. Specifically, we consider a triplet of classes $(a, p, n)$, where $a$ is more visually similar to $p$ than $n$. We assume that by examining the consistency of the neighbourhood of class $a$ in the view of its visual signature, the VAWE of the class $a$ and $p$ should be pulled closer while the VAWE of the class $a$ and $n$ should be pushed far part. Hereafter, we call class $a$, $p$ and $n$ anchor class, positive class and negative class respectively. The training objective is to ensure the distance between $f(\bs_a)$ and $f(\bs_p)$ is smaller than the distance between $f(\bs_a)$ and $f(\bs_n)$. Therefore we employ a triplet hinge loss function:
\begin{align}\label{eq:LOSS}
 \sum_{\forall (\bs_a, \bs_p, \bs_n) \in \calT} \left[  \|f(\bs_a) - f(\bs_p) \|_2^2 - \|f(\bs_a) - f(\bs_n) \|_2^2  + \alpha \right]_+,
\end{align}where $[]_+$ denotes the hinge loss and $\alpha$ is an enforced margin between the distances from anchor class to positive and negative classes. Note that our method does {\bf not} map the semantic word embedding into a shared space with %
visual feature as in many ZSL methods such as DeViSE \cite{Frome2013NIPS_devise}. The mapping function only applies at the semantic domain. We set $\alpha=1.0$ in all experiments.

\subsection{Triplet selection}
\label{sec: triplet}
The choice of the triplet $(a, p, n)$ plays a crucial role in our method. Our method encourages the output embedding to share neighbourhood with visual signatures. Therefore if two classes are close in the visual domain, but distant in the semantic domain, their semantic embeddings should be pulled closer, and vice versa. Specifically, for an anchor class $a \in \calW_s$, if another class $p \in \calW_s$ is within the top-$K_1$ neighbours $N_v(a, K_1)$ in the view of visual domain but not within the top-$K_2$ ($K_2 > K_1$) neighbours $N_s(a, K_2)$ in the view of semantic domain, then $p$ should be pulled closer to $a$ and we include $p$ as a positive class. On the other hand, if another class $n \in \calW_s$ is within the top-$K_1$ neighbours of $a$ in semantic view but not within the top-$K_2$ neighbours in visual view, $n$ should be pushed far away from $a$ and we include $n$ as the negative class. Note that using $K_2 > K_1$ avoids over-sensitive decision on the neighbourhood boundary. In other words, if $j$ is within the top-$K_1$ neighbourhood of $i$, it is deemed ``close'' to $i$ and only if $j$ is not within the top-$K_2$ neighbourhood of $i$, it is considered as ``distant'' from $i$.

As noted by \cite{hubness2015}, nearest neighbour approaches may suffer from the hubness problem in high dimension: some items are similar to all other items and thus become hubs. In our experiment if a positive concept appears in the neighbourhood of many words during training, the VAWE $f(\bs)$ would concentrate around this hub and this could be harmful for learning a proper $f(\cdot)$. We design a simple-but-effective hubness correction mechanism as a necessary regularizor for training by removing such hub vectors from the positive class candidates as the training progresses. We calculate the ``hubness level'' for each concept before each epoch. Concretely,  we accumulate the each concept's times of appearances in the neighbourhood of other concepts in the mapped semantic domain $f(\bs)$. We mark the concepts that appear too often in the neighbourhood of other concepts as hubs and remove them from positive classes in the next training epoch. In our experiment the hubness correction usually brings 2-3\% of improvement over the ordinary triplet selection. We summarize the triplet selection process and hub vectors generation in Algorithm~\ref{ALG:triplet} and Algorithm~\ref{ALG:hubs}, respectively.

\def\ADot{ { $\bf \cdot$ } }%
\begin{minipage}[t]{6cm}
  \vspace{0pt}
  \begin{algorithm}[H]
    \caption{Dynamic triplet selection at epoch $t+1$}
    { \footnotesize
    \KwIn{Nearest neighbourhood structure sets $N_v(i, K)$ and $N_s(i, K)$ in visual and semantic domains for each seen class computed from semantic and visual signatures $(\bs_i, \bv_i), i \in \calW_s$ ; $K_1$ and $K_2$;
    hub vector set at epoch $t$ $H_t$ .
    }
    { Initialize triplet set $\calT_{t+1} =  \varnothing$ at epoch $t+1$.
   }\\
      \For { $i = 1 \cdots | \calW_s |$}
   {
          \ADot
          $\hat{N_v}(i, K) = N_v(i, K) - H_t$, \///remove hubs.\\
          \ADot
          $a = i$, \\
        \For { $s \in N_s(i, K_1)$}
        {
          \If {$s \notin \hat{N_v}(i, K_2)$}
          {\ADot $n = s$,\\
          		\For { $v \in \hat{N_v}(i, K_1)$}
        {
          \If {$v \notin N_s(i, K_2)$}
          {\ADot $p = v$,\\
          \ADot $\calT_{t+1} = \calT_{t+1} \cup (a, p, n)$.}

        }}

        }

   }
   \ADot
    Randomly shuffle the order of triplets in $\calT_{t+1}$.

    \KwOut{
     $\calT_{t+1}$.
}
    }

\label{ALG:triplet}
  \end{algorithm}
\end{minipage}%
\
\begin{minipage}[t]{6cm}
  \vspace{0pt}
  \begin{algorithm}[H]
    \caption{Generating hub vector set before epoch $t+1$}
     { \footnotesize
     \KwIn{output embeddings at epoch $t$ $f_t(\bs_i), i \in \calW_s$;
    number of neighbours in semantic domains $K_1$;}
    { Initialize $Hubs \in \Natural^{| \calW_s |}$ as a zero-valued vector with each of its element counting the hubness level of each vector; hub vector set at epoch $t$ $H_t = \varnothing$.
   }
   \\
    \For { $i = 1 \cdots | \calW_s |$  }
   {
      \ADot
        Get $N_s(i, K_1)$ from $f_t(\bs_i)$, \\
      \For { $j = 1 \cdots | \calW_s |$  }
      {

        \If { $j \in N_s(i, K_1)$}
        { \ADot
        $Hubs_j += 1$,}
      }
   }

   \For { $i = 1 \cdots | \calW_s |$  }
  {

  \If {$Hubs_i > K_1$}
  { \ADot
    $H_t = H_t  \cup  i $
  }
  }
  \KwOut{
     $H_t $.
}
     }
     \label{ALG:hubs}
  \end{algorithm}
\end{minipage}

\subsection{Learning the neural network}
We formulate $f(\cdot)$ as a neural network that takes inputs from the pre-trained word embeddings and outputs new visually aligned word embeddings. During the training stage, the training triplets are selected from the seen classes according to Algorithm~\ref{ALG:triplet}, and parameters of $f(\cdot)$ are adjusted by SGD to minimize the triplet loss \eqref{eq:opt}. Note that although the number of training classes is limited, $f(\cdot)$ is trained with the triplets of classes, which amount up to $\mathcal{O} (| \calW_s |^3)$. The inference structure of $f(\cdot)$ contains two fully-connected hidden layers with ReLU non-linearity, and the output embedding is L-2 normalized to $d'$-D unit hypersphere before being propagated to the triplet loss layer. For a detailed description of the neural network and the training parameters, please refer to the supplementary material.

\begin{align}\label{eq:opt}
& \min_{\Theta} \sum_{\forall (\bs_a, \bs_p, \bs_n) \in \calT_t} \left[  \|f(\bs_a) - f(\bs_p) \|_2^2 - \|f(\bs_a) - f(\bs_n) \|_2^2 + \alpha \right]_+  +  \lambda \| \Theta \|_2^2.
\end{align}
where $\calT_t$ is the set of all selected triplets at epoch $t$.

During the inference stage, $f(\cdot)$ is applied to word embeddings of both seen and unseen classes. The output VAWEs are off-the-shelf for any zero-shot learning tasks.

\section{Experiments}
\label{sec: exp}
In order to conduct a comprehensive evaluation, we train the VAWE from two kinds popular distributed word embeddings: word2vec \cite{word2vec} and GloVe \cite{glove}. We apply the the trained VAWE to four state-of-the-art methods. We compare the performance against the original word embeddings and other ZSL methods using various semantic embeddings. %

\begin{table*}[!t]
\begin{center}
\begin{tabular}{|l|c|c|c|c|c|c|}
\hline
Method &Feature &Embedding &aPY &AwA  &CUB &SUN\\
\hline\hline
Lampert \cite{Lampert14pami} & V &continuous attribute  &38.16 & 57.23 & &72.00 \\
Deng \cite{deng2014large} & D &class hierarchy  & & 44.2 &  & \\
Ba \cite{Ba2015ICCV} & V  &web documents  & &  &12.0  & \\
Akata \cite{Akata15output} &V &word2vec & & 51.2  & 28.4 & \\
Akata \cite{Akata15output} &V &GloVe & & 58.8 &  24.2 & \\
Akata \cite{Akata15output} &V &continuous attribute  & & 66.7 &  50.1 & \\
Qiao \cite{Qiao_2016_CVPR} &V &web documents & &66.46 &29.00 & \\
Zhang \cite{Zhang2015ICCV} &V &continuous attribute &46.23 &76.33 &30.41 &82.50 \\
Zhang \cite{Zhang2016CVPR} &V &continuous attribute &50.35 &79.12 &41.78 &83.83 \\
VSAR \cite{LongBMVC16} &L &continuous attribute &39.42 &51.75 & & \\
SynC \cite{SynC2016} &G & continuous attribute   & &69.7  &53.4 & 62.8  \\
LatEm \cite{LatEm2016} &G &continuous attribute &  &72.5   &45.6 & \\
ESZSL \cite{Romera2015ZSL} &V &continuous attribute &24.22 &75.32  &    &82.10 \\
\hline
\hline

ConSE \cite{NorouziConse13} &V & word2vec  &21.82 & 46.80   & 23.12 & 43.00 \\
ConSE \cite{NorouziConse13} + Ours &V & VAWE word2vec &35.29 & 61.24 &27.44 & 63.10 \\
ConSE \cite{NorouziConse13} &V & GloVe  &35.17 & 51.21   &  & \\
ConSE \cite{NorouziConse13} + Ours &V & VAWE GloVe &42.21 & 59.26 & &  \\
SynC \cite{SynC2016} &V & word2vec   &28.53 &56.71  & 21.54 & 68.00  \\
SynC \cite{SynC2016} + Ours &V & VAWE word2vec  &33.23 & 66.10   & 21.21 & 70.80 \\
SynC \cite{SynC2016} &V& GloVe   &29.92 &60.74  & &   \\
SynC \cite{SynC2016} + Ours &V & VAWE GloVe   &31.88 & 64.51   &  &  \\
LatEm \cite{LatEm2016} &V & word2vec &19.64  & 50.84   & 16.52 & 52.50 \\
LatEm \cite{LatEm2016} + Ours &V & VAWE word2vec &35.64 & 61.46   & 19.12 & 61.30 \\
LatEm \cite{LatEm2016} &V & GloVe &27.72  & 46.12   &  &  \\
LatEm \cite{LatEm2016} + Ours &V & VAWE GloVe &37.29 & 55.51   &  &  \\
ESZSL \cite{Romera2015ZSL} &V & word2vec & 28.32  & 58.12   & 24.82 & 64.50 \\
ESZSL \cite{Romera2015ZSL} + Ours &V &  VAWE word2vec & 43.23  & 76.16 &24.10 & 71.20 \\
ESZSL \cite{Romera2015ZSL} &V & GloVe & 34.53  & 59.72   &  &  \\
ESZSL \cite{Romera2015ZSL} + Ours &V & VAWE GloVe & 44.25  & 75.10 & &  \\
\hline
\end{tabular}
\end{center}
\caption{ZSL classification results on 4 datasets. Blank spaces indicate these methods are not tested on the corresponding datasets. Bottom part: methods using  VAWE and the original word embeddings as semantic embeddings. Upper part: state-of-the-art methods using various sources of semantic embeddings. Visual features include V:VGG-19; G:GoogLeNet; D:DECAF; L:low-level features.}
\label{tab: zsl-all}
\end{table*}

{\bf Datasets:} We test the methods on four widely used benchmark dataset for zero-shot learning: aPascal/aYahoo object dataset \cite{Farhadi09describingobjects} (aPY), Animals with Attributes \cite{Lampert09unseen} (AwA), Caltech-UCSD birds-200-2011 \cite{CUB_200_2011} (CUB), and the SUN scene attribute dataset \cite{SUNDBijcv} (SUN). %

{\bf Distributed word embeddings:} We train the VAWE from two pre-trained distributed word emedding models: word2vec and GloVe. We pre-train the word2vec model from scratch on a large combination of text corpus.
The resulted model generates 1000-D real valued vectors for each concept.  As for GloVe, we use the pre-trained 300-D word embeddings provided by \cite{glove}.
  We only test GloVe on aPY and AwA datasets because the pre-trained GloVe model does not contain the associated word embeddings for too many fine-grained categories in CUB and SUN. %

{\bf Image features and visual signatures:} For all the four test methods in our experiments, we extract the image features from the fully connected layer activations of the deep CNN VGG-19 \cite{Simonyan14VGG}. As aforementioned, we use the average VGG-19 features of each seen category as the visual signatures for them.

{\bf Test ZSL methods:} We apply trained VAWEs on four state-of-the-art methods denoted as ConSE \cite{NorouziConse13}, SynC \cite{SynC2016}, LatEm \cite{LatEm2016} and ESZSL \cite{Romera2015ZSL} in the following sections.%

{\bf Implementation details:} We stop the training when the triplet loss stops decreasing. This usually takes 150 epochs for aPY, 250 epochs for AwA, 50 epochs for CUB and 20 epochs for SUN. Number of nearest neighbours in visual space is $K_1=10$ for all datasets. Number of nearest neighbours in semantic space $K_2$ is set to half the number of seen classes for each dataset (except for SUN, which has many seen classes), that is 10, 20, 75 and 200 for aPY, AwA, CUB and SUN,  respectively. The output dimension is set to $d'$=128. More detailed experimental settings and results are provided in the supplementary material.  %

\subsection{Performance improvement and discussion}
In this section, we test the effect of using VAWE trained from word2vec and GloVe in various ZSL methods. The main results of VAWE compared against the original word embeddings are listed in the bottom part of Table~\ref{tab: zsl-all}. Except for the fine-grained dataset CUB, the VAWEs trained from both word embeddings gain overall performance improvement on all test methods. Most notably on the coarse-grained datasets, i.e., aPY and AwA, the VAWEs outperform their original counterparts by a very large margin.

For ZSL methods, we find that the performance improvement is most significant for ConSE and ESZSL, partly because these two methods directly learns a linear mapping between visual and semantic embeddings. A set of semantic embeddings that is inconsistent with the visual domain would hurt their performance the most. By using the VAWEs, those methods learn a much better aligned visual-semantic mapping and earns a great performance improvement. %

The performance improvement is limited on fine-grained datasets CUB and SUN. Compared to the coarse-grained class datasets, the difference in categories in CUB and SUN is subtle in both visual and semantic domains. This causes the their visual signatures and semantic embeddings more entangled and results higher hubness level. Therefore it is more challenging to re-align the word embeddings of fine-grained categories by our method.

Overall, the VAWEs exhibit consistent performance gain for various methods on various datasets (improved performance on 22 out of 24 experiments). This observation suggests that VAWE is able to serve as a general tool for improving the performance of ZSL approaches.

\subsection{Comparison against the state-of-the-art}
We also compare the improved results of VAWE against the results of recently published state-of-the-art ZSL methods using various sources of semantic embeddings in the upper part of Table~\ref{tab: zsl-all}. It can be observed that methods using VAWE beat all other methods using non-attribute embeddings. Even compared against the best performing attribute-based methods, our results are still very competitive on coarse-grained class datasets: only a small margin lower than \cite{Zhang2016CVPR} that uses continuous attributes. The results indicate that VAWE is a potential substitution for human-labelled attributes. The VAWE is not only human labour free but also provides comparable performance to attributes.

\subsection{The effect of visual features}

\begin{table}[h]
\begin{center}
\begin{tabular}{|l|c|c|c|}
\hline
Visual signature source &Low-level &DeCAF &VGG-19  \\
\hline
\hline
ConSE + Ours &55.57   &60.08 &61.24 \\
SynC + Ours &59.06 &67.30  &66.10  \\
LatEm + Ours &58.43 &63.33 &61.46   \\
ESZSL + Ours &67.28 &73.23 &76.16   \\

\hline
\end{tabular}
\end{center}
\caption{ZSL accuracies on the AwA dataset of VAWE trained with visual signatures from different feature sources. For the ZSL methods, the VGG-19 features are still used for training and testing.}
\label{tab: zsl-sig}
\end{table}

The learning process of the mapping function relies on the choice of visual features which implicitly affects the neighbourhood structure in the visual domain. In this section, we investigate the impact of the choice of visual features on the quality of the mapped VAWE. Again, the quality of the VAWE is measured by its performance on ZSL. In previous sections, we extracted the visual signature as the mean of the VGG-19 features of each class. Here we further replace it with low-level features or DeCAF features provided by \cite{Lampert09unseen} and use them to obtain the VAWEs of word2vec. Once the VAWEs is learned we apply the ZSL with VGG-19 features and the experiment is conducted on the AwA dataset. Note that both DeCAF features and low-level features are weaker image features than VGG-19. The experiment results are shown in Table~\ref{tab: zsl-sig}. From the experimental results, we find that performance of all four ZSL methods do not change too much when we replace VGG-19 with DeCAF to learn the mapping function. Using low-level features will degrade the performance but comparing to the performance of using the original word2vec the learned VAWE still shows superior performance. These observation suggests that we may use one type of visual features to train the VAWE and apply them to ZSL methods trained with another kind of visual features and still obtain good results.

\section{Conclusion}
In this paper, we show that the discrepancy of visual features and semantic embeddings negatively impacts the performance of ZSL approaches. Motivated by that, we propose to learn a neural network with triplet loss to map the word embeddings into a new space in which the neighbourhood structure of the mapped word embedding becomes similar to that in the visual domain. The visually aligned word embeddings boost the ZSL performance to a level that is competitive to human defined attributes. Besides that, our approach is independent of any particular ZSL method. This gives it much more flexibility to generalize to more potential applications of vision-language tasks.

\section*{Acknowledgement}
L. Liu was in part supported by ARC DECRA Fellowship DE170101259.
C. Shen was in part supported by ARC Future Fellowship FT120100969.

\bibliography{CSRef}

\begin{thebibliography}{29}
\providecommand{\natexlab}[1]{#1}
\providecommand{\url}[1]{\texttt{#1}}
\expandafter\ifx\csname urlstyle\endcsname\relax
  \providecommand{\doi}[1]{doi: #1}\else
  \providecommand{\doi}{doi: \begingroup \urlstyle{rm}\Url}\fi

\bibitem[Akata et~al.(2013)Akata, Perronnin, Harchaoui, and
  Schmid]{Akata13label}
Zeynep Akata, Florent Perronnin, Zaid Harchaoui, and Cordelia Schmid.
\newblock Label-embedding for attribute-based classification.
\newblock In \emph{Proc. IEEE Conf. Comp. Vis. Patt. Recogn.}, June 2013.

\bibitem[Akata et~al.(2015)Akata, Reed, Walter, Lee, and
  Schiele]{Akata15output}
Zeynep Akata, Scott Reed, Daniel Walter, Honglak Lee, and Bernt Schiele.
\newblock Evaluation of output embeddings for fine-grained image
  classification.
\newblock In \emph{Proc. IEEE Conf. Comp. Vis. Patt. Recogn.}, June 2015.

\bibitem[Ba et~al.(2015)Ba, Swersky, Fidler, and Salakhutdinov]{Ba2015ICCV}
Jimmy~Lei Ba, Kevin Swersky, Sanja Fidler, and Ruslan Salakhutdinov.
\newblock Predicting deep zero-shot convolutional neural networks using textual
  descriptions.
\newblock In \emph{Proc. IEEE Int. Conf. Comp. Vis.} IEEE, 2015.

\bibitem[Changpinyo et~al.(2016)Changpinyo, Chao, Gong, and Sha]{SynC2016}
Soravit Changpinyo, Wei{-}Lun Chao, Boqing Gong, and Fei Sha.
\newblock Synthesized classifiers for zero-shot learning.
\newblock In \emph{Proc. IEEE Conf. Comp. Vis. Patt. Recogn.}, 2016.

\bibitem[Deng et~al.(2014)Deng, Ding, Jia, Frome, Murphy, Bengio, Li, Neven,
  and Adam]{deng2014large}
Jia Deng, Nan Ding, Yangqing Jia, Andrea Frome, Kevin Murphy, Samy Bengio, Yuan
  Li, Hartmut Neven, and Hartwig Adam.
\newblock Large-scale object classification using label relation graphs.
\newblock In \emph{Proc. Eur. Conf. Comp. Vis.}, 2014.

\bibitem[Dinu and Baroni(2015)]{hubness2015}
Georgiana Dinu and Marco Baroni.
\newblock Improving zero-shot learning by mitigating the hubness problem.
\newblock In \emph{Proc. Int. Conf. Learn. Representations}, 2015.

\bibitem[Farhadi et~al.(2009)Farhadi, Endres, Hoiem, and
  Forsyth]{Farhadi09describingobjects}
Ali Farhadi, Ian Endres, Derek Hoiem, and David Forsyth.
\newblock Describing objects by their attributes.
\newblock In \emph{Proc. IEEE Conf. Comp. Vis. Patt. Recogn.}, 2009.

\bibitem[Frome et~al.(2013)Frome, Corrado, Shlens, Bengio, Dean, Ranzato, and
  Mikolov]{Frome2013NIPS_devise}
Andrea Frome, Greg Corrado, Jon Shlens, Samy Bengio, Jeffrey Dean, Marc'Aurelio
  Ranzato, and Tomas Mikolov.
\newblock {DeViSE}: A deep visual-semantic embedding model.
\newblock In \emph{Proc. Advances in Neural Inf. Process. Syst.}, 2013.

\bibitem[Fu and Sigal(2016)]{fu2016semi}
Yanwei Fu and Leonid Sigal.
\newblock Semi-supervised vocabulary-informed learning.
\newblock In \emph{Proc. IEEE Conf. Comp. Vis. Patt. Recogn.}, 2016.

\bibitem[Jetley et~al.(2015)Jetley, Romera{-}Paredes, Jayasumana, and
  Torr]{SaumyaBMVC15}
Saumya Jetley, Bernardino Romera{-}Paredes, Sadeep Jayasumana, and Philip H.~S.
  Torr.
\newblock Prototypical priors: From improving classification to zero-shot
  learning.
\newblock In \emph{Proc. British Machine Vis. Conf.}, 2015.

\bibitem[Kottur et~al.(2016)Kottur, Vedantam, Moura, and
  Parikh]{Kottur_2016_CVPR}
Satwik Kottur, Ramakrishna Vedantam, Jose M.~F. Moura, and Devi Parikh.
\newblock Visual word2vec (vis-w2v): Learning visually grounded word embeddings
  using abstract scenes.
\newblock In \emph{Proc. IEEE Conf. Comp. Vis. Patt. Recogn.}, June 2016.

\bibitem[Lampert et~al.(2009)Lampert, Nickisch, and Harmeling]{Lampert09unseen}
Christoph~H. Lampert, Hannes Nickisch, and Stefan Harmeling.
\newblock Learning to detect unseen object classes by betweenclass attribute
  transfer.
\newblock In \emph{Proc. IEEE Conf. Comp. Vis. Patt. Recogn.}, 2009.

\bibitem[Lampert et~al.(2014)Lampert, Nickisch, and Harmeling]{Lampert14pami}
Christoph~H. Lampert, Hannes Nickisch, and Stefan Harmeling.
\newblock Attribute-based classification for zero-shot visual object
  categorization.
\newblock \emph{{IEEE} Trans. Pattern Anal. Mach. Intell.}, 2014.

\bibitem[Lazaridou et~al.(2015)Lazaridou, Pham, and
  Baroni]{Lazaridou_combine_vs2015}
Angeliki Lazaridou, Nghia~The Pham, and Marco Baroni.
\newblock Combining language and vision with a multimodal skip-gram model.
\newblock In \emph{NAACL HLT}, 2015.

\bibitem[Li et~al.(2015)Li, Guo, and Schuurmans]{LiICCV15}
X.~Li, Y.~Guo, and D.~Schuurmans.
\newblock Semi-supervised zero-shot classification with label representation
  learning.
\newblock In \emph{Proc. IEEE Int. Conf. Comp. Vis.}, pages 4211--4219, 2015.

\bibitem[Long et~al.(2016)Long, Liu, and Shao]{LongBMVC16}
Yang Long, Li~Liu, and Ling Shao.
\newblock Attribute embedding with visual-semantic ambiguity removal for
  zero-shot learning.
\newblock In \emph{Proc. British Machine Vis. Conf.}, 2016.

\bibitem[Long et~al.(2017)Long, Liu, Shao, Shen, Ding, and Han]{LongCVPR17}
Yang Long, Li~Liu, Ling Shao, Fumin Shen, Guiguang Ding, and Jungong Han.
\newblock From zero-shot learning to conventional supervised classification:
  Unseen visual data synthesis.
\newblock In \emph{Proc. IEEE Conf. Comp. Vis. Patt. Recogn.}, 2017.

\bibitem[Mikolov et~al.(2013)Mikolov, Sutskever, Chen, Corrado, and
  Dean]{word2vec}
Tomas Mikolov, Ilya Sutskever, Kai Chen, Gregory~S. Corrado, and Jeffrey Dean.
\newblock Distributed representations of words and phrases and their
  compositionality.
\newblock In \emph{Proc. Advances in Neural Inf. Process. Syst.}, 2013.

\bibitem[Norouzi et~al.(2014)Norouzi, Mikolov, Bengio, Singer, Shlens, Frome,
  Corrado, and Dean]{NorouziConse13}
Mohammad Norouzi, Tomas Mikolov, Samy Bengio, Yoram Singer, Jonathon Shlens,
  Andrea Frome, Greg Corrado, and Jeffrey Dean.
\newblock Zero-shot learning by convex combination of semantic embeddings.
\newblock In \emph{Proc. Int. Conf. Learn. Representations}, 2014.

\bibitem[Pennington et~al.(2014)Pennington, Socher, and Manning]{glove}
Jeffrey Pennington, Richard Socher, and Christopher~D. Manning.
\newblock {GloVe}: Global vectors for word representation.
\newblock In \emph{Conf. Empirical Methods in Natural Language Processing:
  EMNLP}, 2014.

\bibitem[Qiao et~al.(2016)Qiao, Liu, Shen, and van~den Hengel]{Qiao_2016_CVPR}
Ruizhi Qiao, Lingqiao Liu, Chunhua Shen, and Anton van~den Hengel.
\newblock Less is more: Zero-shot learning from online textual documents with
  noise suppression.
\newblock In \emph{Proc. IEEE Conf. Comp. Vis. Patt. Recogn.}, June 2016.

\bibitem[Romera-Paredes and Torr(2015)]{Romera2015ZSL}
Bernardino Romera-Paredes and Philip~H.S. Torr.
\newblock An embarrassingly simple approach to zero-shot learning.
\newblock \emph{Proc. Int. Conf. Mach. Learn.}, 2015.

\bibitem[Simonyan and Zisserman(2014)]{Simonyan14VGG}
K.~Simonyan and A.~Zisserman.
\newblock Very deep convolutional networks for large-scale image recognition.
\newblock In \emph{Proc. Int. Conf. Learn. Representations}, 2014.

\bibitem[Socher et~al.(2013)Socher, Ganjoo, Manning, and Ng]{SocherNIPS2013}
Richard Socher, Milind Ganjoo, Christopher~D Manning, and Andrew Ng.
\newblock Zero-shot learning through cross-modal transfer.
\newblock In \emph{Proc. Advances in Neural Inf. Process. Syst.}, 2013.

\bibitem[Wah et~al.(2011)Wah, Branson, Welinder, Perona, and
  Belongie]{CUB_200_2011}
C.~Wah, S.~Branson, P.~Welinder, P.~Perona, and S.~Belongie.
\newblock {The Caltech-UCSD Birds-200-2011 Dataset}.
\newblock Technical Report CNS-TR-2011-001, California Institute of Technology,
  2011.

\bibitem[Xian et~al.(2016)Xian, Akata, Sharma, Nguyen, Hein, and
  Schiele]{LatEm2016}
Yongqin Xian, Zeynep Akata, Gaurav Sharma, Quynh Nguyen, Matthias Hein, and
  Bernt Schiele.
\newblock Latent embeddings for zero-shot classification.
\newblock In \emph{Proc. IEEE Conf. Comp. Vis. Patt. Recogn.}, 2016.

\bibitem[Xiao et~al.(2014)Xiao, Ehinger, Hays, Torralba, and Oliva]{SUNDBijcv}
Jianxiong Xiao, Krista~A. Ehinger, James Hays, Antonio Torralba, and Aude
  Oliva.
\newblock Sun database: Exploring a large collection of scene categories.
\newblock \emph{Int. J. Comput. Vision}, 2014.

\bibitem[Zhang and Saligrama(2015)]{Zhang2015ICCV}
Ziming Zhang and Venkatesh Saligrama.
\newblock Zero-shot learning via semantic similarity embedding.
\newblock In \emph{Proc. IEEE Int. Conf. Comp. Vis.} IEEE, 2015.

\bibitem[Zhang and Saligrama(2016)]{Zhang2016CVPR}
Ziming Zhang and Venkatesh Saligrama.
\newblock Zero-shot learning via joint latent similarity embedding.
\newblock In \emph{Proc. IEEE Conf. Comp. Vis. Patt. Recogn.} IEEE, 2016.

\end{thebibliography}
\end{document}